\documentclass[letterpaper, 10 pt, journal, twoside]{ieeetran}
\usepackage{amsmath,amsfonts}
\usepackage{algorithmic}
\usepackage{algorithm}
\usepackage{array}
\usepackage[caption=false,font=normalsize,labelfont=sf,textfont=sf]{subfig}
\usepackage{textcomp}
\usepackage{stfloats}
\usepackage{url}
\usepackage{verbatim}
\usepackage{graphicx}
\usepackage{cite}

\usepackage{subfloat}
\usepackage{subfig}
\usepackage{diagbox}
\usepackage{epstopdf}
\usepackage{pifont}
\usepackage{amsmath}
\usepackage{multirow}
\usepackage{url}
\usepackage{verbatim}
\usepackage{footmisc}
\usepackage{color}
\usepackage{tabularx}
\usepackage{float}
\usepackage{booktabs}
\usepackage{makecell}
\usepackage{bm}

\usepackage{enumitem}
\usepackage{amsmath}
\usepackage{graphicx}

\usepackage{mars}

\begin{document}
\bibliographystyle{IEEEtran}

\title{Robust Lifelong Indoor LiDAR Localization\\ using the Area Graph}


\author{Fujing Xie$^{1}$, S\"{o}ren Schwertfeger$^{1}$%
	\thanks{Manuscript received: June, 16, 2023; Revised September, 25, 2023; Accepted November, 1, 2023.}
	\thanks{This paper was recommended for publication by Editor Ashis Banerjee upon evaluation of the Associate Editor and Reviewers' comments.
		This work has been partially funded by the Shanghai Frontiers Science Center of Human-centered Artificial Intelligence. This work was also supported by the Science and
Technology Commission of Shanghai Municipality (STCSM), project 22JC1410700 ”Evaluation of real-time localization and
mapping algorithms for intelligent robots”.} 
	\thanks{$^{1}$The authors are with the Key Laboratory of Intelligent Perception and Human-Machine Collaboration -- ShanghaiTech University, Ministry of Education, China.
		{\tt\footnotesize \{xiefj, soerensch\}@shanghaitech.edu.cn}}%
	\thanks{Digital Object Identifier (DOI): see top of this page.}
}




\markboth{IEEE Robotics and Automation Letters. Preprint Version. Accepted NOVEMBER, 2023}
{Xie \MakeLowercase{\textit{et al.}}: Robust Lifelong Indoor LiDAR Localization using the Area Graph} 
%
%


\marsPublishedIn{Accepted for:} 		

\marsVenue{IEEE Robotics and Automation Letters (RA-L) 2023}

\marsYear{2023}

\marsPlainAutors{Fujing Xie, S\"{o}ren Schwertfeger}


\marsMakeCitation{Robust Lifelong Indoor LiDAR Localization using the Area Graph}{IEEE Press}

\marsDOI{\url{https://dx.doi.org/10.1109/LRA.2023.3334158}}

\marsIEEE{}


\makeMARStitle
\maketitle
\thispagestyle{empty}
\pagestyle{empty}

\begin{abstract}
%
Lifelong indoor localization in a given map is the basis for navigation of autonomous mobile robots. In this letter, we address the problem of robust localization in cluttered indoor environments like office spaces and corridors using 3D LiDAR point clouds in a given Area Graph, which is a hierarchical, topometric semantic map representation that uses polygons to demark areas such as rooms, corridors or buildings. This representation is very compact, can represent different floors of buildings through its hierarchy and provides semantic information that helps with localization, like poses of doors and glass. In contrast to this, commonly used map representations, such as occupancy grid maps or point clouds, lack these features and require frequent updates in response to environmental changes (e.g. moved furniture), unlike our approach{\color{black}, AGLoc}, which matches against lifelong architectural features such as walls and doors. For that we apply filtering to remove clutter from the 3D input point cloud and then employ further scoring and weight functions for localization. Given a broad initial guess from WiFi 
{\color{black} and barometer} 
localization, our experiments show that our global localization and the weighted point to line ICP pose tracking perform very well, even when compared to localization and SLAM algorithms that use the current, feature-rich cluttered map for localization.




\end{abstract}


\section{Introduction}
\IEEEPARstart{R}{obust}, efficient, and accurate localization serves as the foundation for higher functionalities in intelligent autonomous robotics, including navigation, decision making, and task execution. There are several ways to achieve localization for autonomous robots, such as
 GPS. But GPS does not work indoors. Some applications use markers, but we prefer a system that does not require modifications of the environment. Most common localization approaches then utilize a particle filter or scan matching against some form of map. 

The most common map representations utilized for indoor localization are 2D occupancy grid maps, 3D point clouds and sets of visual information, such as bag of words. All of these have significant disadvantages when it comes to robust, lifelong indoor localization. {\color{black}Our map representation has three key aspects:} (1) A very compact representation, such that huge environments like a campus or district can be easily stored on a robot or transferred over the Internet. (2)~The representation should be stable over time, i.e. not requiring updates, thus mapping permanent structures of buildings, like walls and doors. Visual approaches struggle with this, as they are easily distracted by clutter, especially since walls often are with low texture. (3) Additionally, it would be great if the map used for localization is easy to acquire, e.g. not requiring a mapping robot to perform Simultaneous Localization and Mapping (SLAM). This could be achieved by creating said maps from architectural CAD data, which is usually readily available for every building. In this letter we propose to utilize our hierarchical, topometric, semantic map representation Area Graph for localization with 3D LiDAR data.

\begin{figure}[t]
	\centering
	\includegraphics[width=0.40\textwidth]{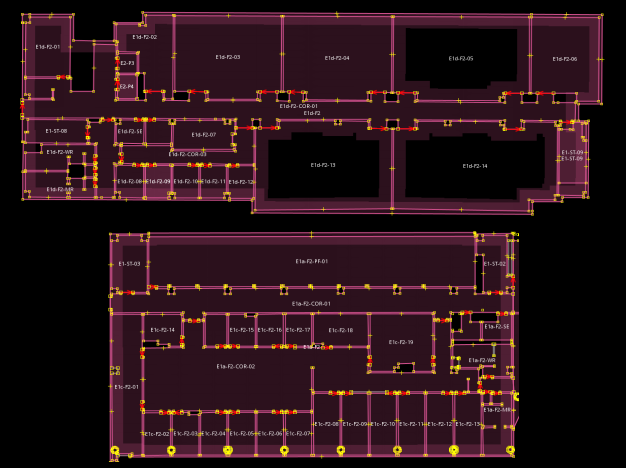}
	\caption{Open Street Map Area Graph maps for the two experimental datasets Seq01 (top) and Seq02. We can see areas demarked by pink polygons and passages (doors) as red lines, that are connections between areas and semantic information in terms of room names and passages. Yellow dots are polygon coordinates which form areas and passages.}
	\label{fig:area_graph}
	\vspace{-7mm}
\end{figure}

\begin{figure*}[t]
	\centering
	\begin{center} 
		\includegraphics[width=0.82\textwidth]{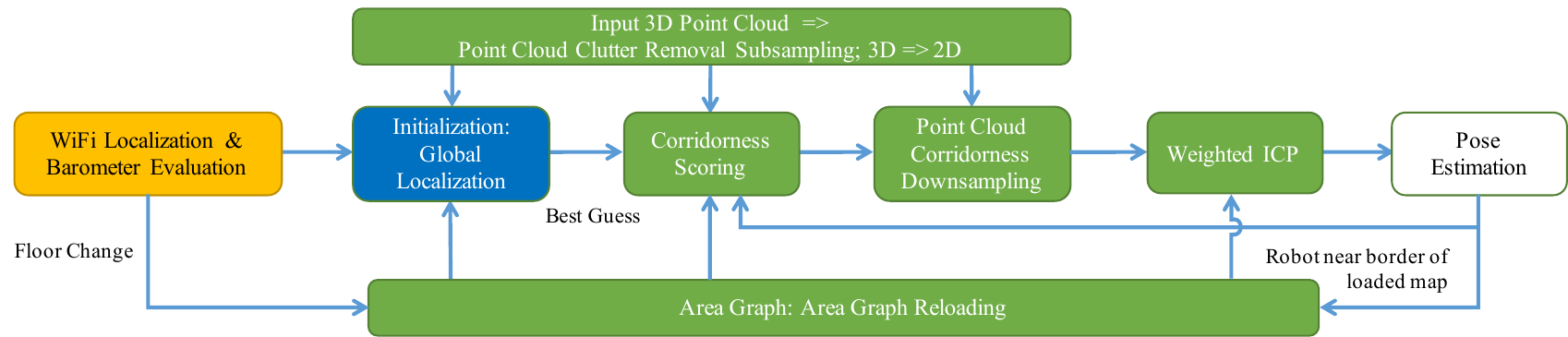}
		\caption{The overall framework of 
			{\color{black}AGLoc}.}
		\label{fig:overview}
	\end{center}
	\vspace{-7.5mm}
\end{figure*}

The topometric 2D Area Graph has been introduced in \cite{hou2019area, hou2022area}, utilizing areas defined by polygons as graph nodes, which represent physical spaces in the environment, such as rooms and corridors. The edges of the graph are so-called passages, defined as line segments of the area polygons, that connect two neighboring areas. Following the ideas presented in \cite{he2021hierarchical}, we have extended the {\color{black}Area Graph }with a hierarchical representation, where parent areas enclose their children to represent bigger units, such as floors or whole buildings. This way we now also represent 3D information such as floor level and elevation. We store this data as tags in the open street map osm XML file format\cite{haklay2008openstreetmap}, which we call osmAG. For brevity, the details of  osmAG are omitted here, since our Area Graph localization mainly relies on the polygon data of the leaf areas {\color{black}(leaf of the hierarchical tree)} of a building floor and their passages. Fig. \ref{fig:area_graph} shows topometric and semantic data of one floor of a building, that is used in Seq01  and Seq02 in our experiments.

The main motivation for developing our Area Graph LiDAR localization{\color{black}, that we dubbed AGLoc, }is its intended use in a navigation system that will enable robots to autonomously drive from any room in a campus to any other. {\color{black}We proposed the global and local planning with osmAG in \cite{feng2023osmag}, and will} concentrate on indoor localization here. {\color{black}The goal of our work is to provide robust lifelong localization, not to outperform traditional localization approaches that use detailed maps containing clutter.} Outdoors we plan to use GPS.
%
WiFi localization is another indoor localization approach, but its error is quite significant, typically several meters\cite{liu2020survey}. But we expect WiFi localization to be available in almost all indoor places, so we utilize it to confine our search space for global localization. 

{\color{black}AGLoc} proposes a two-stage localization approach. {\color{black}Firstly, we globally localize the robot in the Area Graph. The WiFi and barometer }localization restricts our initial search space. We then uniformly sample pose guesses within that search space and score them by matching a 2D \textit{clutter free point set}, which was subsampled from the initial 3D LiDAR point cloud, with the Area Graph polygons using a special score function. This 2D  \textit{clutter free point set} follows ideas similar to  \cite{he2019furniturefree} to sample points on walls or doors and ignore all other readings, such as on furniture. After obtaining the initial global pose we perform pose tracking, without needing odometry nor IMU data. Here, we utilize the \textit{clutter free point set} in a point to line weighted Iterative Closest Point (ICP) algorithm that is employing a special weight function to further reduce the influence of clutter on the result and which matches against the polygons of the Area Graph. Additionally, we utilize  corridorness downsampling to improve the ICP result in corridor-like environments. 

Our experiments will show that 
{\color{black}AGLoc} achieves accurate localization in both heavily cluttered environments and challenging long corridors, as well as achieving accurate global localization. 
Our contributions are as follows:
\begin{itemize}[=6mm]
\item[$\bullet$] We propose a 3D LiDAR based longterm indoor localization 
{\color{black}using} the Area Graph, both globally and locally.
	\item [$\bullet$] We propose a clutter removal subsample method to remove most clutter.
	\item [$\bullet$] 
	{\color{black}To localize the robot globally, we employ a sampling approach to generate guesses, and then develop a score function to score each guess during global localization, which evaluates their match to the Area Graph.}
	\item [$\bullet$] We propose a weight function along with point to line ICP to ignore clutter, which enables localization in a representation that only uses walls and passages{\color{black}, thus enabling robust lifelong localization even in the presence of non-static obstacles (e.g. furniture, humans).}
	\item [$\bullet$] We present a corridorness score function with which we downsample point clouds for better ICP localization in corridor-like environments.
\end{itemize} 

\vspace{-0.35cm} 
\section{Related Works}

{\color{black}Lidar Odometry and Mapping (LOAM)} based SLAM algorithms\cite{zhang2017low,shan2018lego,shan2020lio} assume no prior knowledge about the environment, localize themselves and build point cloud maps at the same time. They need the robot to map the environment before localizing in it, and these data-rich maps need to be updated whenever the environment changes. There are also several papers on localizing robots in CAD maps or floor plans, e.g., \cite{gao2022fp}\cite{boniardi2017robust}, which extract grid maps from Building Information Models, while the Area Graph uses polygons.

With the development of deep learning, the use of semantic information for localization and mapping is widely adapted. \cite{zimmerman2022long} uses RGB images to perform object detection and add semantic information like table, sink or whiteboard to CAD floor plans, and then uses a 2D LiDAR combine that semantic cues in a Monte Carlo Localization (MCL) framework \cite{dellaert1999MCL}. Cui et al. \cite{cui2021monte} and Zimmerman et al.\cite{zimmerman2022robust} use AI to detect texts and combine textual cues into MCL framework.

Nguyen et al.\cite{nguyen2021autonomous} use learning based classifiers to detect corridors from 2D LiDAR measurements in order to solve the kidnapped robot problem. Hornung et al. \cite{hornung_octomap_2013} use principal component analysis (PCA) on LiDAR scan contours as a corridor detector.
In our approach, with the help of the Area Graph we can easily detect which beams hit the the corridor's dominant direction to compensate for this biased data.

Other approaches utilize map matching on topometric maps, such as CAD data\cite{delin2023floorplannet, hou2022area, schwertfeger2016map, mielle2016using}. The downside here is, that the robot first has to create a grid map with several rooms before localization can be successful. {\color{black}There is also research on localization using Topometric maps such as \cite{mazuran2018relative} and \cite{badino2011visual}}.

The most similiar approch to ours is given by Gao et al.\cite{gao2022fp}. They propose a novel data structure in the form of an
approximate nearest neighbour field (ANNF), which is generated from CAD data
and enables an efficient look-up of the nearest geometric
floor plan elements (e.g. wall segments) for any given
point. It then projects wall points onto the horizontal plane, and uses ICP to register the point cloud from LiDAR measurement with the point cloud generate from the CAD floor plan. However, their work is different and we believe outperformed by our approach in three regards:
First, they sample points on line and curve segments of the CAD floor plan, which is unnecessary in Area Graph since the Area Graph represents an area with a set of sequential
nodes, therefore the look up step is unnecessary.
Second, they assume they have a relatively accurate initial guess, and ignore the global localization.
Third, they don't consider clutter or furniture in the environment.
\vspace{-0.15cm} 
\section{Approach}
Fig. \ref{fig:overview} provides a comprehensive overview of the entire pipeline. The first step is the initial global localization. Afterwards we perform pose tracking. 
{\color{black} Inspired by\cite{ito2014w} and \cite{bisio2018wifi}, the }initial broad localization estimate comes from WiFi {\color{black} and barometer (for the floor of the building - compared against a base station)} localization, with several meters of error.

Using this information we will load the Area Graph (Section \ref{subsec:areagraph}). 
We remove the ceiling and ground from the robot 3D LiDAR point cloud, since we are performing a 2D localization, on a given floor.
The point cloud will then be subsampled to 2D, keeping only furthest points of each column, creating the \textit{clutter free point set}  (Section \ref{subsec:downsampling}). 
The calculation of the intersection of each point's ray with the Area Graph is presented in Section \ref{subsec:intersection}. 
Finally, the initial localization is computed by calculating scores for a set of pose guesses uniformly sampled around the initial broad WiFi {\color{black}and barometer }localization, where the best scored pose is selected as the global localization result (Section \ref{subsec:scoring}).

The pose tracking is also using the \textit{clutter free point set} with a weighted point to line ICP algorithm\cite{censi2008icp}. The lines are the polygons from the Area Graph. The initial guess comes initially from the global localization (see above), and subsequently from the previous tracking result, without having to rely on odometry nor IMU data. The weight function for the ICP is introduced in Section \ref{section:weight function} and the weighted ICP algorithm is detailed in Section \ref{subsec:icp}. 
We also propose an optimization in which we 
further subsample the 2D \textit{clutter free point set}, that is used in the ICP algorithm, using a \textit{Corridorness Scoring} function, which improves the ICP performance in corridor-like environments (Section \ref{subsec:corridorness}). 

\vspace{-0.15cm} 
\subsection{Area Graph Loading and Reloading}
\label{subsec:areagraph}


Through the hierarchical structure of the Area Graph, the space can be partitioned into various spatial divisions, such as regions, areas (e.g. a university campus), buildings, floors in buildings, and so on. For global localization we utilize WiFi {\color{black}and barometer} localization, which we assume to be present in all environments that our algorithm should be employed in. WiFi has an error of several meters (e.g. we assume a radius of 6m in our experiments), so we need to employ our approach to 1) figure out which area (e.g. room, corridor) we are in and 2) where precisely we are in that area. 
{\color{black}Comparing the air pressure via a robot-mounted barometer with a base-station barometer we can reliable estimate the floor we are in, using the metric height information stored in the osmAG.
With the WiFi localization we can get all areas of the current floor. }
The localization approach presented in this paper relies only on the accurate metric information of the polygons of walls and passages (doors). Therefore, we are only interested in the leafs of that floor of the hierarchical osmAG, thus a detailed description of the osmAG is {\color{black}omitted here but presented in \cite{feng2023osmag}}. This letter makes no use of the topological information stored in the osmAG, the hieracial properties of osmAG only come into play w.r.t. loading leaf areas of the current floor, semantic information is used for the identification of passages (doors) and the most important aspect of the osmAG is the metric information about the position of the points in the polygons (which are converted from {\color{black}longitude, latitude} format to Cartesian x, y coordinates).

%
Whenever the robot transits between floors using an elevator/ stairs or reaches the boundary of the currently loaded Area Graph, it becomes necessary to reload the Area Graph specific to the corresponding floor and location.

\vspace{-0.20cm} 
\subsection{Point Cloud Clutter Removal Subsampling}
\label{subsec:downsampling}

Both the global localization as well as the pose tracking utilize the step presented in this section to generate the 2D \textit{clutter free point set}, which is then matched against the Area Graph. The Area Graph is representing walls (and passages/ doors as special cases) as 2D polygons, therefore the sensor data (3D point cloud) should also be filtered for clutter such as furniture, people and so on. Ideally, the \textit{clutter free point set} should then contain only points on walls or closed doors. The assumption for this Clutter Removal Subsampling is, that every column {\color{black}(set of 64 points from the LiDAR taken at the same time, that have the same azimuth angle, thus forming a vertical column of points)} of the 3D point cloud will have some beams hitting the wall, which should also be the furthest points from the origin for that column. Thus we simply add the 2D coordinates of the furthest point of each column to the  2D \textit{clutter free point set}. That assumption will not always be met, but the hope is that there will only be few such points and that those can be handled by the weight function introduced in Section \ref{section:weight function}. 

Specifically, one frame of LiDAR scan can be seen as a size $[N, M]$ $point_{i,j}$ matrix, where $N$ represents how many rings the LiDAR has (64 in our case), and $M$ means how many points each ring has, which is 600 beams in our case. The LiDAR point cloud is divided into 600 columns, based on the yaw angles of the points (w.r.t. to the LiDAR), with each ring contributing one point to each column. The video accompanying this letter nicely depicts this step. With $\|. , .\|$ as the Euclidean 2D distance (ignoring the z-coordinate) of $point_{i,j}$ from the sensor origin, the  \textit{clutter free point set} $P_{cf}$  is:\begin{equation} 
	\color{black}
 P_{cf} = \{ point_{i^*, j} \,|\, i^* = \arg\max_{i} \|point_{i,j}\|  \} _{j=1}^{M}
\end{equation}

\begin{figure}[b]
	\vspace{-5mm}
	\centering
	\includegraphics[width=0.48\textwidth]{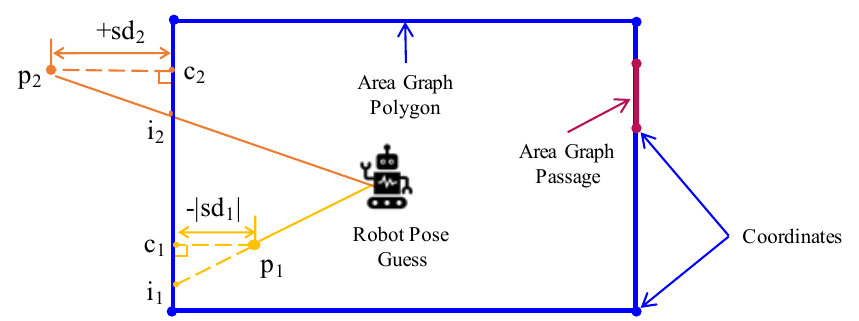}
	\caption{
		Illustration of Area Graph polygons and signed distance $sd_j$ intersection calculation.
	}
	\label{fig:signed_dis}
\end{figure}

Additionally to removing clutter from the point cloud and projecting it from 3D to 2D, this step is also saving significant amounts of computation time, since the number of points to be processed in the following steps is heavily subsampled (in our case $\frac{1}{64}$).

\subsection{Intersection Calculation}
\label{subsec:intersection}

Given the Area Graph, the robot pose guess, and the \textit{clutter free point set}, we can 
calculate the intersection of each point's ray ($\overrightarrow{\mbox{origin point}}$) with Area Graph polygons. In fact, we are mainly interested in the signed distance $sd_j$ from the point $p_j$ to the closest point on the line segment of the polygon $c_j$ intersecting with the ray of that point (positive values are outside the room, negative inside):

\begin{equation}
	\begin{aligned}	
		sd_j &= \| p_j, c_j\|  \\
		sd_j &= -sd_j  \quad  \text{if }   \| p_j \| < \| i_j\|
	\end{aligned}
\end{equation}

See  Fig. \ref{fig:signed_dis} for an illustration. 
In order to speed up the computation, we identify the area of the Area Graph within which the LiDAR scan originates {\color{black}($v_h$)}, because rays should always intersect with its polygon. Passages of the Area Graph can represent doors, which can be open or closed during localization, so special handling is needed. Our approach does this by checking if the signed distance $sd_j$ is bigger than the threshold $th_{passage}$ ( $th_{passage} = 0.1m$ in our experiments). If this is the case, we compute the closest intersection of that ray with any loaded Area Graph polygon, which we assume to be the wall seen through that passage. We use the same approach for polygons which have been marked as being transparent or "can look over" in the semantic information of the Area Graph (e.g. glass railing). 


\begin{figure}[b]
	\vspace{-5mm}
	\centering
	\includegraphics[width=0.30\textwidth]{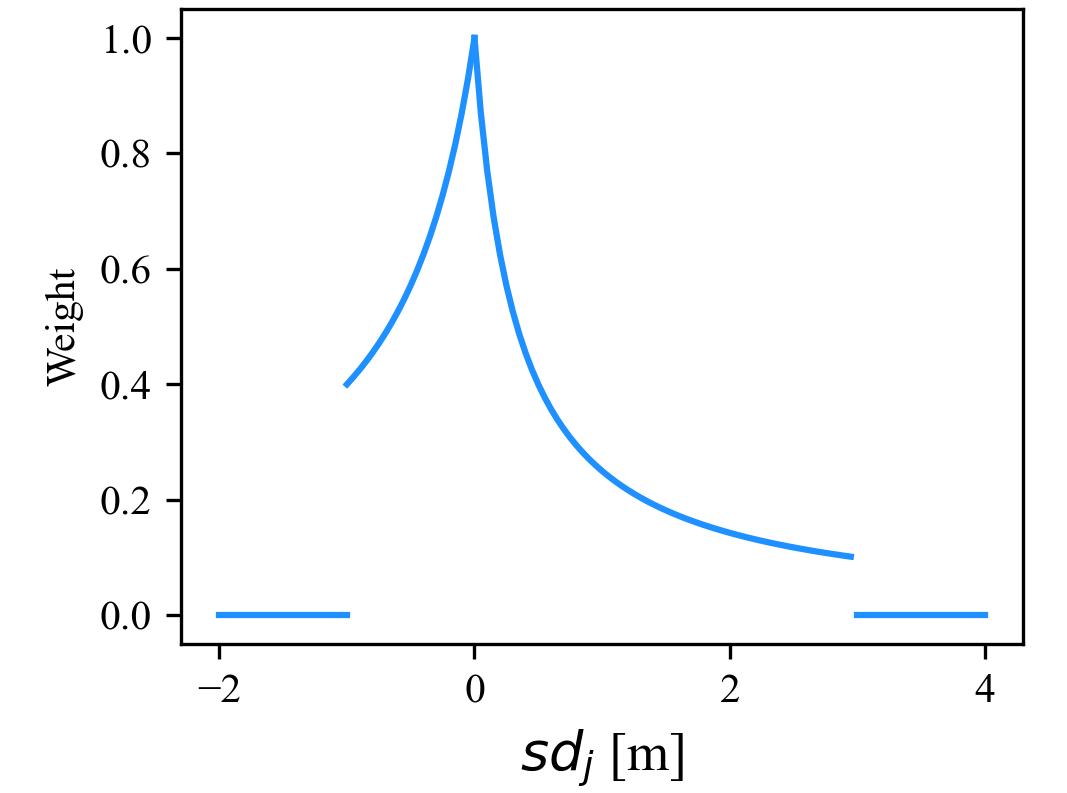}
	\vspace{-3mm}
	\caption{
		Weight function $\mathcal{W}(sd_j)$, computing a weight for the weighted point to line ICP based on the minimum signed distance (positive values behind the wall/ outside) between a point and the line its beam is intersecting with. Mostly ignoring clutter inside rooms (early cut-off of negative $sd_j$) and also ignoring long error beams, e.g. due to reflection.
	}
	\label{fig:weight}
	\vspace{2mm}
\end{figure}

\subsection{Guess Scoring}
\label{subsec:scoring}
The robot's pose is determined by 2D coordinates $p=(x,y)$ and the yaw angle $\theta\in[0,2\pi)$. 
According to the localization from WiFi {\color{black}and barometer}, we sample $n$ guesses in range of $r$ meters radius with a certain distance and 
orientation interval, along with the knowledge of which area $v_h$ each guess is in:
\begin{equation}
	\begin{aligned}	
		&G=\{g_h\}_{h=1}^n  \\
		&g_h=(x_h,y_h,\theta_h,v_h)
	\end{aligned}
	\label{eq:area_graph}
\end{equation}


We employ three error functions to assess the similarity between each guess's transformed \textit{clutter free point set} and the Area Graph polygons. 
We formulate  four different score functions combining the three error functions. In our experimental evaluation in Section \ref{subsec:res_global} we test them all and found equations $S_1$ and $S_3$ to both deliver excellent results and finally selected $S_1$ for our algorithm{\color{black}, for its faster computation speed.}
For each pose guess we calculate the according  intersection distances $sd_j$ with the Area Graph polygons.  We also employ $\mathcal{W}$ as the weight function that is introduced in Section \ref{section:weight function}.

Our three error functions are:
\begin{enumerate}
	\item $nearby\_error$: We use a small threshold $d$ ($0.8m$ in our experiments) to only include points that are quite close to the intersection. Less error means more points are inside the threshold, means a better match, because the error used for points exceeding the threshold $d$ is quite high (2m).

	\begin{equation}
		E_{1}= \sum_{j=1}^{M}
		\begin{cases}
			\| sd_j \| &  \text{if } \| sd_j \| < d  \\
			2 & \text{otherwise}
		\end{cases}
	\end{equation}
	
	\item {\color{black}$weighted\_outside\_fit$}: We ignore points contained within this guess's area (i.e. $sd_j < 0$; see yellow point in Fig. \ref{fig:signed_dis}) since they are more likely hitting clutter (e.g. furniture, cabinets on the wall), and we use a weight function as illustrated in Section \ref{section:weight function} and Fig. \ref{fig:weight} to ignore points that are too far away. 
	
	\begin{equation}
		E_{2}= \sum_{j=1}^{M} 
		\begin{cases}
			\mathcal{W}( sd_j ) &  \text{if } sd_j > 0 \\
			0 & \text{otherwise}
		\end{cases}
	\end{equation}

	\item $unweighted\_outside\_error$: without the weight function of {\color{black}$weighted\_outside\_fit$}.
\end{enumerate}

\begin{equation}
	E_{3}= \sum_{j=1}^{M}
	\begin{cases}
		sd_j &  \text{if } sd_j > 0  \\
		0 & \text{otherwise}
	\end{cases}
\end{equation}

Our four score functions are as follows:

\begin{equation}
	\begin{aligned}
		S_1&=1/E_1 &
		S_2&=E_2\\
		S_3&=E_2/E_1 &
		S_4&=1/E_3\\
	\end{aligned}		
\end{equation}


	%
		%
			
			We choose the guess with the highest score as our global localization result.

			\subsection{Weight Function for Guess Scoring and Weighted ICP}
			\label{section:weight function}

%
%

			We incorporate $\mathcal{W}$ as weight function, 


			\begin{equation}
				\label{eq:weight_function}
	\mathcal{W}(sd_j)=
	\begin{cases}
		0 &  \text{if } sd_j \leq -1 \\
		\frac{1}{1.5\times {\color{black}|}sd_j{\color{black}|}+1}&  \text{if } -1<sd_j \leq 0 \\
		\color{black}\frac{1}{3\times sd_j+1}		&  \text{if } 0<sd_j < 3 \\
		0&  \text{if } sd_j \geq 3 \\
	\end{cases}
\end{equation}

			depicted in Fig. \ref{fig:weight}, for three main reasons: 
			\begin{enumerate}[leftmargin=3.2mm]
				\item Ignore clutter: in heavily cluttered environments, beams that hit inside areas {\color{black}($sd_j \leq 0$)} may encounter clutter, despite employing our clutter removal subsampling from Section \ref{subsec:downsampling}, therefore we assign a smaller threshold {\color{black}($sd_j \leq -1$)} to reject points that are far away from the supposed intersection $i_j$ with the Area Graph.
				\item 	Beams longer than the distance to the wall (Area Graph polygon) that they are intersecting with {\color{black}($sd_j > 0$)}, are most likely a strong indication of a wrong pose guess. Therefore we want to include their error in the score function, to ensure that this guess receives a bad score. 
				On the other hand, very long outside beams {\color{black}($sd_j \geq 3$)} may be due to reflection of the beams (e.g. on shiny metal surfaces or windows), therefore we cut-off outside beams over a certain threshold, as they might otherwise skew the result to an incorrect value. 
				
				\item If there are points close to the polygon we want to reward those. Therefore the function peaks super-linearly around 0.

			\end{enumerate}
		
		{\color{black}According to our tests the chosen parameters in Eq. \ref{eq:weight_function} work well and the localization is not very sensitive to their values. }

				\subsection{Pose Tracking via Weighted Point to Line ICP}
				\label{subsec:icp}
				In this stage, we have a relatively accurate initial guess, either from global localization or the previous pose tracking estimate.  
				Our registration method is a variation of the weighted Iterative Closest
				Point (ICP) algorithm\cite{maurer1996registration} and \cite{bergstrom2014robust}, combined with the idea of point to line ICP from \cite{censi2008icp}.

			We define the registration as 
			\begin{equation}
				\sigma= \mathop{\min}_{R,t}\mathop{\sum}_{j=1}^{M}  \mathcal{W}(\pi(p_j, R, t))\times\|  \pi(p_j, R, t)\|^2
				\label{eq:2}
			\end{equation}
			where $\pi(p_j)$ calculates the corresponding $sd_j$ from the Area Graph and the transformed current point $p_j$ from the {\color{black}2D} \textit{clutter free point set}, as mentioned in Section \ref{subsec:intersection}. $R\in \mathbb{R}^{2\times2}$ is the rotation matrix and $t\in \mathbb{R}^2$ is the translation. $\mathcal{W}$ is the weight function introduced in Section \ref{section:weight function}.
			%

			Our experiments will show that our weigthed point to line ICP approach for pose tracking works very well for localization within an Area Graph map. In the next section we present a nice enhancement of our approach that improves localization performance in corridor-like environments.

\begin{figure}[t]
	\centering
	\includegraphics[width=0.46\textwidth]{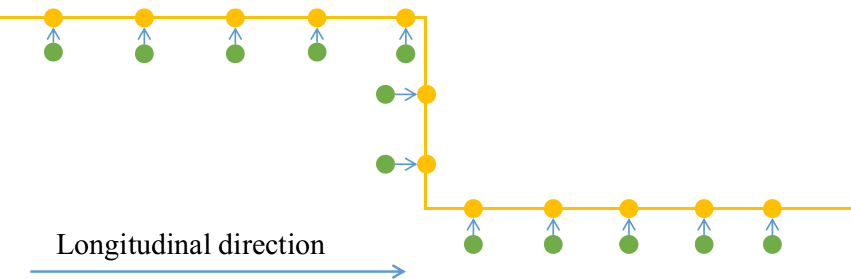}
	\caption{Illustration of problematic ICP performance in corridor-like environments. Only few (2) points match against the vertical line, but many on horizontal lines, fixing the translation in the longitudinal direction in place, preventing the correct solution. Our corridorness downsampling reduces the number of points on lines in the dominant orientation, mitigating this problem.
	}
	\label{fig:point_match}
		\vspace{-5mm}
\end{figure}

\begin{figure}[b]
		\vspace{-7mm}
	\centering
	{\includegraphics[width=1.44in]{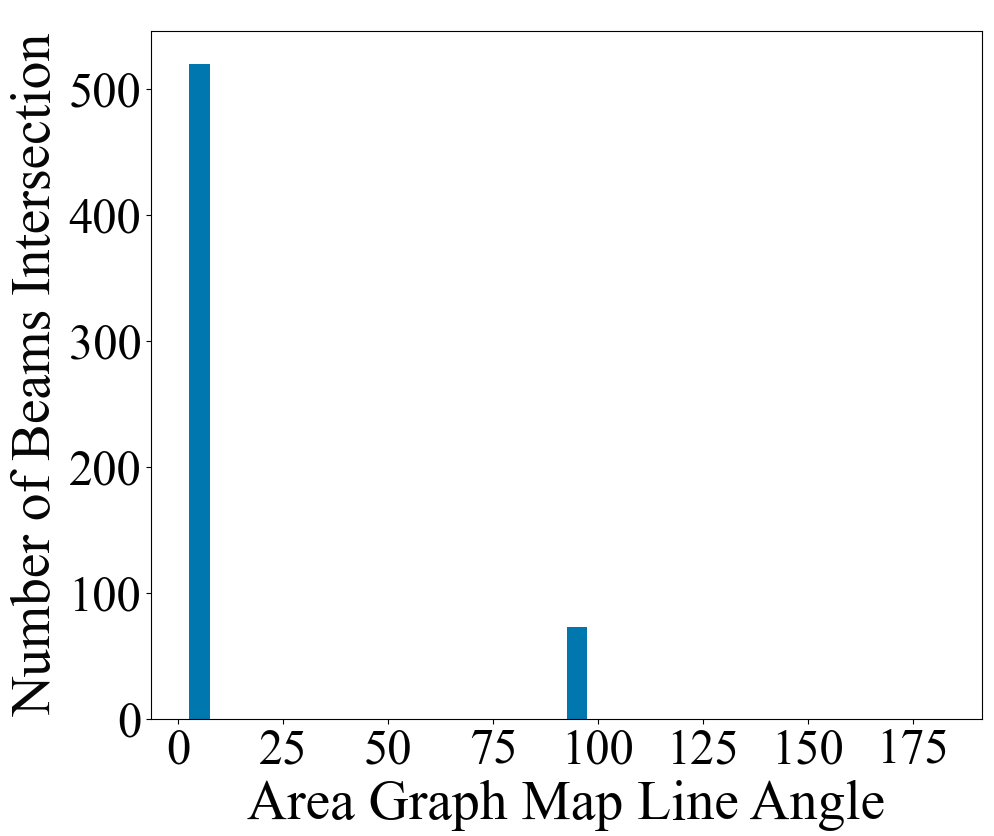}} 
	{\includegraphics[width=1.46in]{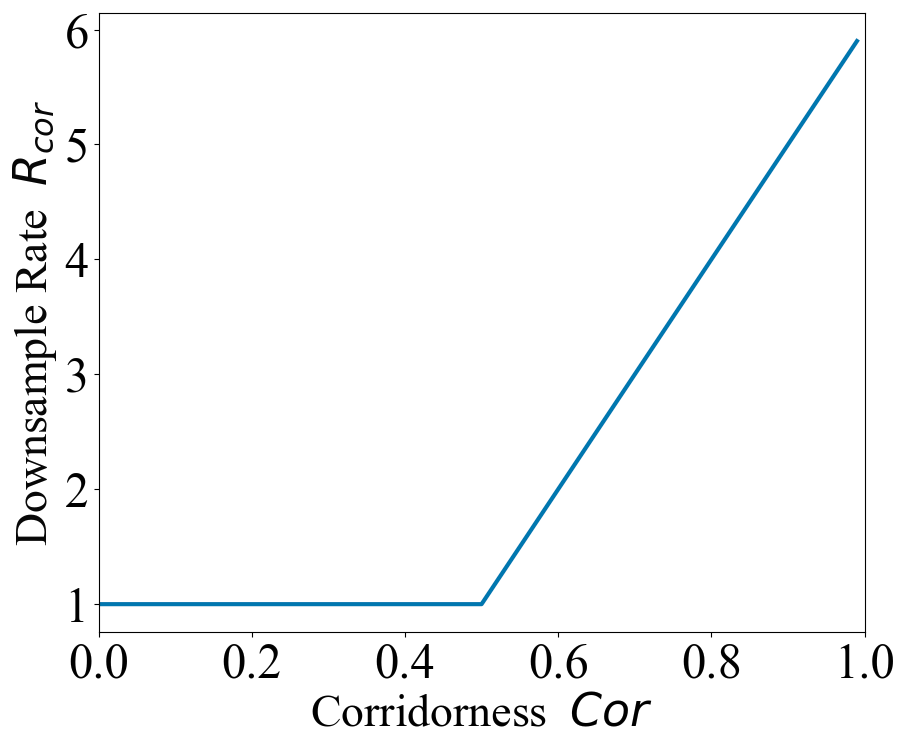}}
	\caption{Corridorness Calculation. (left) Example Histogram for calculating the Corridorness value $Cor$ in a corridor. (right) Function to calculate the downsample rate $R_{cor}$.} 
	\label{fig:histogram} 
	\vspace{-0mm}
\end{figure}

			\subsection{Corridorness Downsampling }  
			\label{subsec:corridorness}
		
		A property of ICP registration, even point to line registration, is, that during the iteration, pairs of points are used. In the common case, sensor points will be matched with map points that are located on various, non-parallel surfaces, which will still move the new pose estimate towards the true robot pose during the closed-form estimation of $R, t$. 
		
		But there are special circumstances (see Fig. \ref{fig:point_match}) in which the scan is predominantly hitting one or more obstacles that are parallel, e.g. in a long corridor. The longitudinal position of the robot within the corridor is not well observed if there are no perturbing features on the corridor walls. In this case ICP will match sensor points with points on the corridor wall at the initial guess. {\color{black}The closed form registration estimate (e.g. via SVD or Horn's Algorithm) that utilizes those point pairs will hold the robot position in place,} even though they could easily slide along the wall, since there are only few points on non-parallel surfaces that would pull the solution towards the true value. 
		This degradation of ICP performance can happen in long corridors, but also in other cases, for example if a robot is close to one wall in a heavily cluttered room. Then our approach would filter out the clutter points and only use the points on this one wall for localization, leading to similarly degradated ICP performance.
		
		As a solution for this problem we further subsample the 2D \textit{clutter free point set} to de-emphasize the dominant orientation. In turn, this emphasizes the non-dominant orientations, which helps ICP to converge to the correct pose. For that we always calculate a corridoness value $Cor$ of the current scan's intersection with the Area Graph, that grows towards $1$ the more points are on lines of the same orientation. 
		
		In detail, we are constructing a histogram with line orientations of the Area Graph as bins ($5^\circ$ interval $\Rightarrow$ 36 bins){\color{black}, which are incremented  for every point matching with these lines} (see Fig. \ref{fig:histogram}). Note, that this step is easy for us to do, because we have the polygon-based Area Graph map representation - doing this on grid maps or in point clouds would be much harder. The corridoness value is then defined as: $Cor=N_{max}/N_{m}$, with $N_{max}$ as the number of points on the most populated bin {\color{black}(dominant orientation)} and  $ N_{m}$ as the total number of entries. In the example of a normal room, half of the sensor points would fall on walls of one orientation, and the other half on walls perpendicular to the first orientation, resulting in a corridoress value $Cor=0.5$, while a scan that only sees one straight wall would get the value $Cor=1.0$.


			We then define a downsample rate $R_{cor}$, that will be used only on points from the  \textit{clutter free point set} that are matched with lines of the maximum bin:
			\begin{equation}
				\label{eq:area_graph}
				R_{cor}=
				\begin{cases}
					1&  \text{if } 0 <Cor \leq 0.5 \\
					10\times Cor-4&  \text{if } 0.5<Cor \leq1 \\
					
				\end{cases}
		\end{equation}
		
		In typical environments there will be no downsampling, since the downsample rate $R_{cor}$, only raises above 1 if there is a dominant orientation. Our experiments will show the effectiveness of our corridorness downsampling. 
		
		A really nice feature of our approach is, that it is using the orientation of the polygon segments of the Area Graph that are matched with a sensor scan. {\color{black}We thus do not need to perform line detection on the LiDAR scan, which could be imprecise.}
		{\color{black}A different method to detect such corridor-like environments would be to add semantic information into the Area Graph of how "corridorness" an area is. }
		%
		 %
		 But our approach is preferable, since it uses the actual intersection of LiDAR scans with the map, and thus inherently considers the actual information encoded in the scan, which may be affected, for example, by a heavily cluttered environment or the maximum range of the sensor. As long as the initial guess places the robot {\color{black}within the correct Area Graph area,} our corridorness downsampling works well, since it mainly relies on the information in the map, which is assumed to be correct.

		%
		%

\section{Experiments}
We test our algorithm in two stages: (1) by utilizing individual LiDAR frames to evaluate the global localization, and (2) by testing the pose tracking outcome against ground truth poses. Additionally, we perform three ablation studies on the pose tracking: (I) for the weight function, (II) the usage of semantic information about passages and (III) the corridoness downsampling. Furthermore, we compare our approach with {\color{black}Adaptive Monte Carlo Localization (AMCL)} using a detailed grid map, a grid map generated from the Area Graph and AMCL without odometry. Finally, we perform a comparison of our approach against the localization of a {\color{black}full fledged} 3D SLAM algorithm. 

\subsection{Datasets and Implementation}
\begin{figure}[t]
	\centering
	\includegraphics[width=0.49\textwidth]{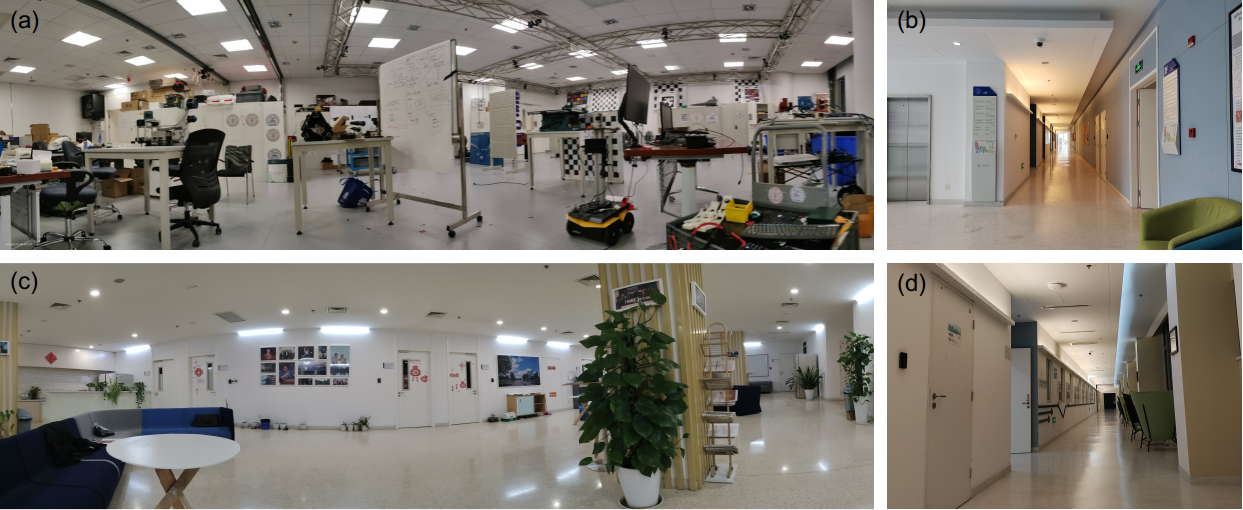}
	\caption{Test environments: (a) Seq01: Our cluttered lab; (b) Seq01: Long corridor; (c) Seq02: Office area; (d) Seq02: Another long corridor}
	\label{fig:lab}
		\vspace{-3mm}
\end{figure}

We collected two datasets, Seq01 and Seq02, of very cluttered rooms and long corridors, as shown in Fig. \ref{fig:lab}. For the corridorness experiment we select subsequences of Seq01 {\color{black} and Seq02} that concentrate on corridors (Fig. \ref{fig:corridor_ape}). The 3D point clouds are captured with a Hesai PandarQT 64 line LiDAR with a vertical field of view (fov) of more than $104^\circ$ and $360^\circ$ horizontal fov, running at 10Hz. The datasets, osmAG and ground truth paths are available online\footnote{\url{https://robotics.shanghaitech.edu.cn/datasets/osmAGlocalization}}. A video of global localization and pose tracking with Seq01 is provided alongside this paper. The source code of our algorithm will be published as open source at a later point as part of the complete osmAG navigation stack. Our algorithm is implemented in C++ and integrated with ROS. We performed all tests on a laptop with an Intel Core i5-9300H CPU @ 2.40GHz and 24GB RAM. We use LIO-SAM's \cite{shan2020lio} trajectory as our ground truth, as LOAM is generally accepted to be a very accurate SLAM algorithm
\cite{elhousni2020survey}, and LIO-SAM is more accurate than our comparison LeGO-LOAM in indoor environments\cite{chan2021lidar}.

\subsection{Global Localization Experiment}
\label{subsec:res_global}
In Section \ref{subsec:scoring} we introduced four scoring functions for global localization, which we test for 507 different poses, each with an associated 3D LiDAR point cloud. This data was taken from Seq01 and Seq02 at one second intervals. As outlined in Section \ref{subsec:scoring}, we sample pose guesses around a WiFi {\color{black}and barometer} localization initial position. In our experiment we add a random error of max 0.25m to the true position (as per our ground truth) and sample with 6m radius with a step size of 0.5m and an angular resolution of 2$^{\circ}$, leading to approx. 80~000 pose guesses.

Firstly, we select which guess score equation leads to the best localization results by checking the success rate of each. Table \ref{table:guess score function_} shows the results when requiring less than 10$^{\circ}$ rotation error and 0.5m (1.0m) translation error. We observed that all maximum translation guesses also fall within our rotation bracket. 

%

We see that both $S_1$ and $S_3$ exhibit a similar success rate, with approximately 80\% of the frames successfully finding $g_h$ within a 0.5m error and a success rate $\geq 98\%$ for a still good error of max 1m. The average computation time for each LiDAR scan's global localization is about 88 seconds. 


\begin{table}[h]
	\vspace{-3mm}
	\caption{{Comparison of Different Score Functions}}
	\label{table:guess score function_}		
	\centering
	\scalebox{0.99}{
	\begin{tabular}{ccccc}
		\toprule
		\multirow{2}{1.2cm}{\shortstack{ Score   \\   Function   }}
		& \multicolumn{2}{c}{\shortstack{Rotation \\Success Rate\\max $10^\circ$ error}}  &\multicolumn{2}{c}{\shortstack{Translation \& Rotation\\Success Rate\\max 0.5m (1.0m) error}} \\ \cmidrule{2-5} 
		&Seq01&Seq02&Seq01&Seq02\\
		\midrule
		$\mathbf{S_1}$ &0.97&0.99& 0.81 (0.98)&0.91 (0.99)  \\
		$S_2$ &0.96&0.98& 0.80 (0.94)& 0.88 (0.96)\\
		$S_3$& 0.97& 1.0& 0.81 (0.97)& 0.92 (0.99)  \\
		$S_4$& 0.77& 0.85& 0.51 (0.68)& 0.60 (0.66)  \\
		
		\bottomrule
	\end{tabular}
}
	\vspace{-7mm}
\end{table}

\begin{figure}[tb]
	\centering
		\includegraphics[width=0.21\textwidth]{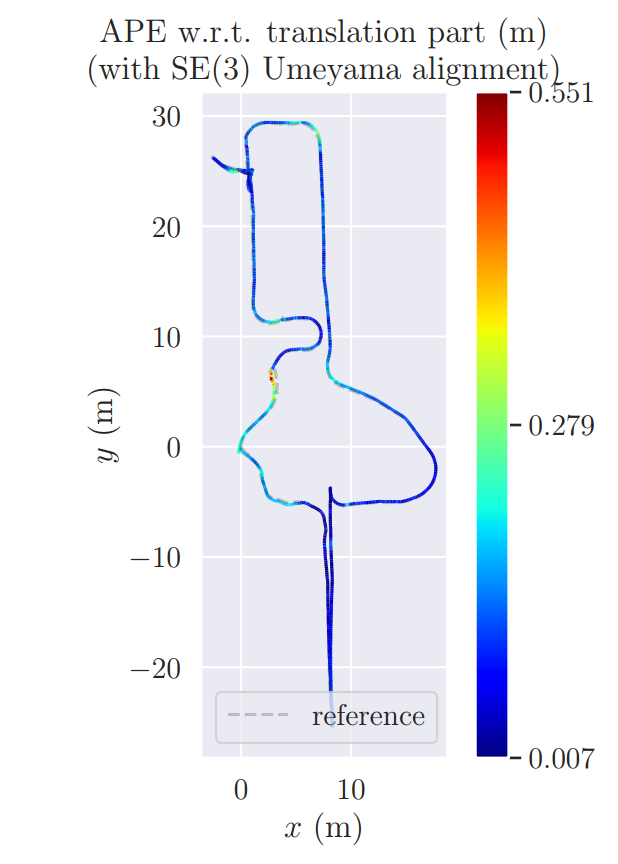}
		\includegraphics[width=0.204\textwidth]{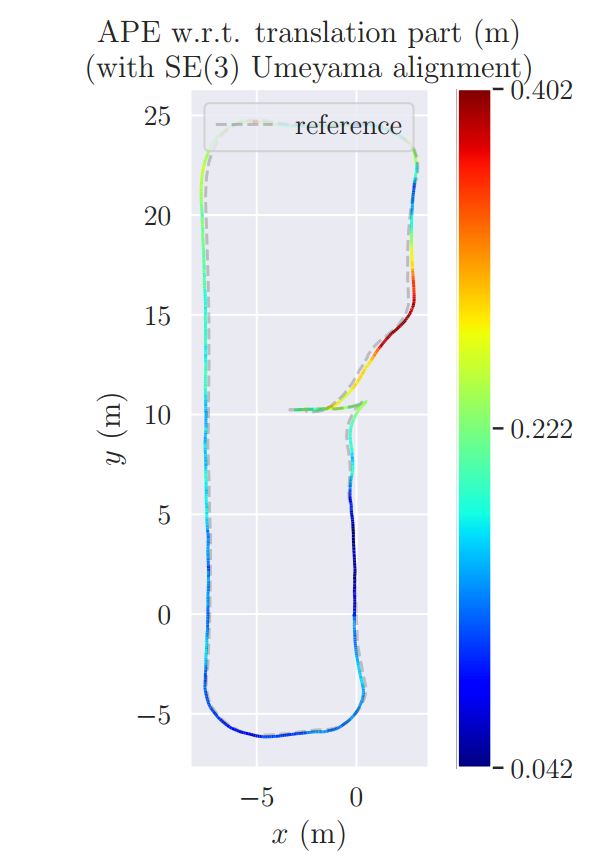}
	\caption{Absolute Pose Error (APE) of Seq01 (left) and Seq02.}
	\label{fig:traj}
\vspace{-6mm}
\end{figure}

\begin{table*}[h]
	\caption{{Pose Tracking Performance Comparison {\color{black}of AGLoc}}}		
	\label{table:pose tracking result}
	\centering
	\scalebox{0.89}{
		\begin{tabular}{lccccccc}
			\toprule
			\multicolumn{2}{c}{\multirow{2}{*}{Method} }&\multirow{2}{*}{Environment}&\multirow{2}{*}{\shortstack{Run Time \\ {[ms]}}} &\multicolumn{2}{c}{ ATE Max[m]} & \multicolumn{2}{c}{ ATE RMSE[m]} \\\cmidrule{5-8} 
			
			&&&&Seq01&Seq02&Seq01&Seq02 \\
			
			\midrule
			
			&\textbf{Ours}&\multirow{5}{*}{Whole Loop}&\multirow{5}{*}{12.5}&0.55&0.40&0.14&0.19\\
			&No Corridorness DS&&&0.47&0.46&0.15&0.21\\
			Our Method:&No Weight Function for ICP&&&0.65&/&0.17&/\\
			
			
			{\color{black}AGLoc}&All Passages Open &&&0.56&0.57&0.14&0.20\\
			&All Passages Close &&&/&/&/&/\\ 
			
			\midrule
			
			\multirow{3}{*}{AMCL}&With Occupancy Map &\multirow{3}{*}{Whole Loop}&\multirow{3}{*}{17.2}&0.38&0.47&0.11&0.25\\
			&With Area Graph &&&0.68&0.47&0.21&0.20\\
			&Only Point Cloud &&&/&/&/&/\\ 
			
			\midrule
			
			LeGO-LOAM&No IMU&Whole Loop&71.0&0.39&0.25&0.093&0.10\\
			
			\midrule \midrule
			
			
			Our Method:&Corridorness DS&\multirow{2}{*}{Corridor}&\multirow{2}{*}{12.5}&0.31&0.45&0.10&0.12\\
			{\color{black}AGLoc}&No Corridorness DS&&&0.39&0.50&0.12&0.15\\ 
			\midrule

			 \multirow{2}{*}{\color{black}AMCL}&\color{black}With Occupancy Map &\multirow{2}{*}{\color{black}Corridor}&\multirow{2}{*}{\color{black}17.2}&\color{black}0.26&\color{black}0.50&\color{black}0.07&\color{black}0.20\\
			&\color{black}With Area Graph &&&\color{black}0.37&\color{black}0.44&\color{black}0.14&\color{black}0.19\\

			\midrule
			\color{black}LeGO-LOAM&\color{black}No IMU&\color{black}Corridor&\color{black}71.0&\color{black}0.37&\color{black}0.22&\color{black}0.09&\color{black}0.09\\
			
			\bottomrule
			\vspace{-8.8mm}
			
		\end{tabular}
	}
\end{table*}


%

\subsection{Pose Tracking Experiments}

The results of our pose tracking experiments are summarized in Table \ref{table:pose tracking result} in terms of run time {\color{black}as well as} maximum and RMSE (Root Mean Squared Error) of ATE (Absolute Trajectory Error). ATE is calculated using evo \cite{grupp2017evo} against the ground truth poses as provided by LIO-SAM. Fig.~\ref{fig:traj} depicts the trajectory errors of both sequences.

\begin{figure}[b]
	\vspace{-7mm}
	\centering
	\includegraphics[width=0.40\textwidth]{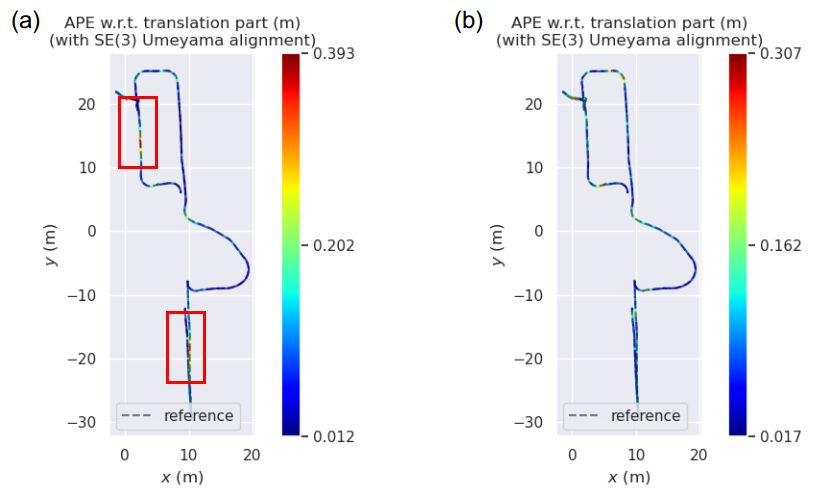}
	\caption{Comparison with and without corridorness {\color{black}scoring}, (a) Pose tracking without corridorness scoring, large error in corridors compared to (b); (b) Pose tracking with corridorness scoring with less error inside corridors.}
	\label{fig:corridor_ape}
\end{figure}

\subsubsection{Comparison with AMCL}
AMCL is a widely use 2D LiDAR localization solution using occupancy grid maps and odometry, which may include IMU estimates. Actually, to give AMCL the best opportunity to perform very well, we utilize LIO-SAM\cite{shan2020lio}'s IMU odometry as the odometry source for the AMCL prediction phase, which offers superior accuracy compared to other odometry methods.  Firstly, we test the AMCL performance with a 2D grid map generated by gmapping from the same dataset. Secondly, we provide AMCL with a rendered grid map version of the Area Graph. The third AMCL experiment uses no  LIO-SAM odometry to highlight how well our approach works even though it does not utilize any odometry. Instead, the odometry estimates for AMCL are provided using the ROS (Robot operating system) package 'laser\_scan\_matcher'. That experiment has the same result when using gmapped and Area Graph map: fail.

In Table \ref{table:pose tracking result} AMCL demonstrates a slight performance advantage over our method when using the gmapping grid map. We believe that this is mainly due to the fact that all the clutter is in the map and thus provides valid data for the scan for map matching, as well as due to the excellent odometry data. Running AMCL with the rendered Area Graph performs worse than our approach. Still, its results are not too bad, which again may be partly attributed to the odometry data. Using AMCL with the  'laser\_scan\_matcher' loses localization rapidly. {\color{black} AMCL in the corridor environment performs worse than ours when utilizing the Area Graph map and has mixed results with the occupancy map. }

\subsubsection{Comparison with LeGO-LOAM}
We also compared our 2D localization with the popular 3D SLAM method LeGO-LOAM, which yielded better results. However, it should be noted that LeGO-LOAM consumes more time. Additionally, since it does not have a pure localization mode, it is not really suitable for a navigation stack in which we need drift-free, longterm localization in a given map. 

	%
	%
	%
	%

\subsubsection{Ablation Studies}
Table  \ref{table:pose tracking result} also shows ablation studies. First, we evaluate the effect of the corridorness downsampling. Especially the experiment in the corridor environment demonstrates superior performance when utilizing the corridorness score to downsample the point cloud, as 
depicted in Fig. \ref{fig:corridor_ape}.

Next, we run our algorithm with ICP where all weights are fixed to 1 ("No Weight Function for ICP"), showing that the weight function is needed to further improve localization performance by filtering clutter and error points. Without it Seq02 even fails. 

Finally, we show the importance of utilizing the semantic information in the Area Graph about doors (passages) and transparent material (also marked as passages in our experiments - later we will introduce a dedicated semantic tag for it). The results show that considering all doors open all the time has slightly worse results than our approach, which adapts to both cases. Treating all doors as closed is not working at all, especially when being close to a door.

\section{Conclusion and Discussion}

This paper presents robust lifelong indoor LiDAR global localization and pose tracking using the hierarchical, semantic topometric Open Street Map Area Graph map representation. One key aspect of our algorithm is the clutter removal subsampling of the 3D LiDAR point clouds and subsequent matching with the Area Graph. We calculate score values of several thousand pose guesses around the WiFi {\color{black} and barometer} initial guess, whose maximum then determines the global localization. 
Our pose tracking employs a weighted point to line ICP, where the weights come from a special {\color{black}clutter-removal function.} A corridorness downsampling further improves ICP performance in corridor-like environments.

Our experiments and ablation studies show the excellent performance of our localization approach. Relying on very sparse, abstract and small map data, {\color{black}we successfully globally localize the robot with the help of WiFi localization} and can perform accurate and robust pose tracking, without the need for IMU nor odometry data. Given this good performance, we feel that it is not necessary to incorporate our method into a Monte Carlo Localization approach. We show that our method is not affected by clutter in the environment, but instead localizes against architectural features such as walls and doors, and thus claims to be a lifelong localization method. We use semantic information (doors, transparent material) and line orientation for corridorness detection to further improve the robustness of our approach. The initial global localization is relatively slow (88 seconds), but we hope to improve this speed through implementation and algorithm improvements in the future. The pose tracking takes 12.5ms per frame, faster than the algorithms we compared against, also thanks to the sparse map representation. We are working on integrating this Area Graph localization into our Area Graph navigation stack to enable autonomy for mobile robots in campus-scale environments\cite{feng2023osmag}. {\color{black} That stack will also integrate odometry and IMU measurements for even better localization. }

Our approach is relying on the dense, 360 degree LiDAR  data our sensor provides. Even though RGBD sensors are providing dense point clouds, they wouldn't work with our algorithm, as they only see one wall or corner, which is not sufficient for our approach. Older LiDAR sensors such as the Velodyne Puck have fewer number of beams and narrower vertical fields of view and would often not be able to see wall points. But LiDAR technology has developed tremendously in recent years, delivering more dense data with greater field of view and falling sensor prices, so our approach is now very feasible for practical autonomous robots.

\bibliography{Bibliography}

\begin{thebibliography}{10}
\providecommand{\url}[1]{#1}
\csname url@samestyle\endcsname
\providecommand{\newblock}{\relax}
\providecommand{\bibinfo}[2]{#2}
\providecommand{\BIBentrySTDinterwordspacing}{\spaceskip=0pt\relax}
\providecommand{\BIBentryALTinterwordstretchfactor}{4}
\providecommand{\BIBentryALTinterwordspacing}{\spaceskip=\fontdimen2\font plus
\BIBentryALTinterwordstretchfactor\fontdimen3\font minus
  \fontdimen4\font\relax}
\providecommand{\BIBforeignlanguage}[2]{{%
\expandafter\ifx\csname l@#1\endcsname\relax
\typeout{** WARNING: IEEEtran.bst: No hyphenation pattern has been}%
\typeout{** loaded for the language `#1'. Using the pattern for}%
\typeout{** the default language instead.}%
\else
\language=\csname l@#1\endcsname
\fi
#2}}
\providecommand{\BIBdecl}{\relax}
\BIBdecl

\bibitem{hou2019area}
J.~Hou, Y.~Yuan, and S.~Schwertfeger, ``Area graph: Generation of topological
  maps using the voronoi diagram,'' in \emph{2019 19th International Conference
  on Advanced Robotics (ICAR)}.\hskip 1em plus 0.5em minus 0.4em\relax IEEE,
  2019, pp. 509--515.

\bibitem{hou2022area}
J.~Hou, Y.~Yuan, Z.~He, and S.~Schwertfeger, ``Matching maps based on the area
  graph,'' \emph{Intelligent Service Robotics}, 2022.

\bibitem{he2021hierarchical}
Z.~He, H.~Sun, J.~Hou, Y.~Ha, and S.~Schwertfeger, ``Hierarchical topometric
  representation of 3d robotic maps,'' \emph{Autonomous Robots}, vol.~45,
  no.~5, pp. 755--771, 2021.

\bibitem{haklay2008openstreetmap}
M.~Haklay and P.~Weber, ``Openstreetmap: User-generated street maps,''
  \emph{IEEE Pervasive computing}, vol.~7, no.~4, pp. 12--18, 2008.

\bibitem{feng2023osmag}
D.~Feng, C.~Li, Y.~Zhang, C.~Yu, and S.~Schwertfeger, ``osmag: Hierarchical
  semantic topometric area graph maps in the osm format for mobile robotics,''
  \emph{arXiv preprint arXiv:2309.04791}, 2023.

\bibitem{liu2020survey}
F.~Liu, J.~Liu, Y.~Yin, W.~Wang, D.~Hu, P.~Chen, and Q.~Niu, ``Survey on
  wifi-based indoor positioning techniques,'' \emph{IET communications},
  vol.~14, no.~9, pp. 1372--1383, 2020.

\bibitem{he2019furniturefree}
Z.~He, J.~Hou, and S.~Schwertfeger, ``Furniture free mapping using 3d lidars,''
  in \emph{2019 IEEE International Conference on Robotics and Biomimetics
  (ROBIO)}, IEEE.\hskip 1em plus 0.5em minus 0.4em\relax IEEE, 2019.

\bibitem{zhang2017low}
J.~Zhang and S.~Singh, ``Low-drift and real-time lidar odometry and mapping,''
  \emph{Autonomous Robots}, vol.~41, pp. 401--416, 2017.

\bibitem{shan2018lego}
T.~Shan and B.~Englot, ``Lego-loam: Lightweight and ground-optimized lidar
  odometry and mapping on variable terrain,'' in \emph{2018 IEEE/RSJ
  International Conference on Intelligent Robots and Systems (IROS)}.\hskip 1em
  plus 0.5em minus 0.4em\relax IEEE, 2018, pp. 4758--4765.

\bibitem{shan2020lio}
T.~Shan, B.~Englot, D.~Meyers, W.~Wang, C.~Ratti, and D.~Rus, ``Lio-sam:
  Tightly-coupled lidar inertial odometry via smoothing and mapping,'' in
  \emph{2020 IEEE/RSJ international conference on intelligent robots and
  systems (IROS)}.\hskip 1em plus 0.5em minus 0.4em\relax IEEE, 2020, pp.
  5135--5142.

\bibitem{gao2022fp}
L.~Gao and L.~Kneip, ``Fp-loc: Lightweight and drift-free floor plan-assisted
  lidar localization,'' in \emph{2022 International Conference on Robotics and
  Automation (ICRA)}.\hskip 1em plus 0.5em minus 0.4em\relax IEEE, 2022, pp.
  4142--4148.

\bibitem{boniardi2017robust}
F.~Boniardi, T.~Caselitz, R.~K{\"u}mmerle, and W.~Burgard, ``Robust lidar-based
  localization in architectural floor plans,'' in \emph{2017 IEEE/RSJ
  International Conference on Intelligent Robots and Systems (IROS)}.\hskip 1em
  plus 0.5em minus 0.4em\relax IEEE, 2017, pp. 3318--3324.

\bibitem{zimmerman2022long}
N.~Zimmerman, T.~Guadagnino, X.~Chen, J.~Behley, and C.~Stachniss, ``Long-term
  localization using semantic cues in floor plan maps,'' \emph{IEEE Robotics
  and Automation Letters}, vol.~8, no.~1, pp. 176--183, 2022.

\bibitem{dellaert1999MCL}
F.~Dellaert, D.~Fox, W.~Burgard, and S.~Thrun, ``Monte carlo localization for
  mobile robots,'' in \emph{Proceedings 1999 IEEE International Conference on
  Robotics and Automation}, vol.~2, 1999, pp. 1322--1328 vol.2.

\bibitem{cui2021monte}
L.~Cui, C.~Rong, J.~Huang, A.~Rosendo, and L.~Kneip, ``Monte-carlo localization
  in underground parking lots using parking slot numbers,'' in \emph{2021
  IEEE/RSJ International Conference on Intelligent Robots and Systems
  (IROS)}.\hskip 1em plus 0.5em minus 0.4em\relax IEEE, 2021, pp. 2267--2274.

\bibitem{zimmerman2022robust}
N.~Zimmerman, L.~Wiesmann, T.~Guadagnino, T.~L{\"a}be, J.~Behley, and
  C.~Stachniss, ``Robust onboard localization in changing environments
  exploiting text spotting,'' in \emph{2022 IEEE/RSJ International Conference
  on Intelligent Robots and Systems (IROS)}.\hskip 1em plus 0.5em minus
  0.4em\relax IEEE, 2022, pp. 917--924.

\bibitem{nguyen2021autonomous}
P.~T.-T. Nguyen, S.-W. Yan, J.-F. Liao, and C.-H. Kuo, ``Autonomous mobile
  robot navigation in sparse lidar feature environments,'' \emph{Applied
  Sciences}, vol.~11, no.~13, p. 5963, 2021.

\bibitem{hornung_octomap_2013}
A.~Hornung, K.~M. Wurm, M.~Bennewitz, C.~Stachniss, and W.~Burgard,
  ``\BIBforeignlanguage{en}{{OctoMap}: an efficient probabilistic {3D} mapping
  framework based on octrees},'' \emph{\BIBforeignlanguage{en}{Autonomous
  Robots}}, vol.~34, no.~3, pp. 189--206, Apr. 2013.

\bibitem{delin2023floorplannet}
D.~Feng, Z.~He, J.~Hou, S.~Schwertfeger, and L.~Zhang, ``Floorplannet: Learning
  topometric floorplan matching for robot localization,'' in \emph{IEEE
  International Conference on Robotics and Automation}.\hskip 1em plus 0.5em
  minus 0.4em\relax IEEE, 2023.

\bibitem{schwertfeger2016map}
S.~Schwertfeger and A.~Birk, ``Map evaluation using matched topology graphs,''
  \emph{Autonomous Robots}, vol.~40, pp. 761--787, 2016.

\bibitem{mielle2016using}
M.~Mielle, M.~Magnusson, and A.~J. Lilienthal, ``Using sketch-maps for robot
  navigation: Interpretation and matching,'' in \emph{2016 IEEE international
  symposium on safety, security, and rescue robotics (SSRR)}.\hskip 1em plus
  0.5em minus 0.4em\relax IEEE, 2016, pp. 252--257.

\bibitem{mazuran2018relative}
M.~Mazuran, F.~Boniardi, W.~Burgard, and G.~D. Tipaldi, ``Relative topometric
  localization in globally inconsistent maps,'' \emph{Robotics Research: Volume
  2}, pp. 435--451, 2018.

\bibitem{badino2011visual}
H.~Badino, D.~Huber, and T.~Kanade, ``Visual topometric localization,'' in
  \emph{2011 IEEE Intelligent vehicles symposium (IV)}.\hskip 1em plus 0.5em
  minus 0.4em\relax IEEE, 2011, pp. 794--799.

\bibitem{ito2014w}
S.~Ito, F.~Endres, M.~Kuderer, G.~D. Tipaldi, C.~Stachniss, and W.~Burgard,
  ``W-rgb-d: floor-plan-based indoor global localization using a depth camera
  and wifi,'' in \emph{2014 IEEE international conference on robotics and
  automation (ICRA)}.\hskip 1em plus 0.5em minus 0.4em\relax IEEE, 2014, pp.
  417--422.

\bibitem{bisio2018wifi}
I.~Bisio, A.~Sciarrone, L.~Bedogni, and L.~Bononi, ``Wifi meets barometer:
  Smartphone-based 3d indoor positioning method,'' in \emph{2018 IEEE
  International Conference on Communications}.\hskip 1em plus 0.5em minus
  0.4em\relax IEEE, 2018, pp. 1--6.

\bibitem{censi2008icp}
A.~Censi, ``An icp variant using a point-to-line metric,'' in \emph{2008 IEEE
  International Conference on Robotics and Automation}.\hskip 1em plus 0.5em
  minus 0.4em\relax Ieee, 2008, pp. 19--25.

\bibitem{maurer1996registration}
C.~R. Maurer, G.~B. Aboutanos, B.~M. Dawant, R.~J. Maciunas, and J.~M.
  Fitzpatrick, ``Registration of 3-d images using weighted geometrical
  features,'' \emph{IEEE transactions on medical imaging}, vol.~15, no.~6, pp.
  836--849, 1996.

\bibitem{bergstrom2014robust}
P.~Bergstr{\"o}m and O.~Edlund, ``Robust registration of point sets using
  iteratively reweighted least squares,'' \emph{Computational optimization and
  applications}, vol.~58, pp. 543--561, 2014.

\bibitem{elhousni2020survey}
M.~Elhousni and X.~Huang, ``A survey on 3d lidar localization for autonomous
  vehicles,'' in \emph{2020 IEEE Intelligent Vehicles Symposium (IV)}.\hskip
  1em plus 0.5em minus 0.4em\relax IEEE, 2020, pp. 1879--1884.

\bibitem{chan2021lidar}
T.~H. Chan, H.~Hesse, and S.~G. Ho, ``Lidar-based 3d slam for indoor mapping,''
  in \emph{2021 7th International Conference on Control, Automation and
  Robotics (ICCAR)}.\hskip 1em plus 0.5em minus 0.4em\relax IEEE, 2021, pp.
  285--289.

\bibitem{grupp2017evo}
M.~Grupp, ``evo: Python package for the evaluation of odometry and slam.''
  \url{https://github.com/MichaelGrupp/evo}, 2017.

\end{thebibliography}
\vfill

\end{document}